\documentclass[runningheads]{llncs}
\usepackage{graphicx}

\begin{document}
\title{A transformer based approach for fighting COVID-19 fake news}

\author{S.M. Sadiq-Ur-Rahman Shifath\inst{1}\orcidID{0000-0003-2428-6595} \and
Mohammad Faiyaz Khan\inst{2}\orcidID{0000-0002-2155-5991} \and
Md. Saiful Islam\inst{3}\orcidID{0000-0001-9236-380X}}
\authorrunning{S.M. S.R. Shifath et al.}
%
\institute{Department of Computer Science and Engineering,\\ Shahjalal University of Science and Technology, Sylhet, Bangladesh
\email{shifathrahman472533@gmail.com}\and
\email{mfaiyazkhan@student.sust.edu}\and
\email{saiful-cse@sust.edu}}

\maketitle
\begin{abstract}
The rapid outbreak of COVID-19 has caused humanity to come to a stand-still and brought with it a plethora of other problems. COVID-19 is the first pandemic in history when humanity is the most technologically advanced and relies heavily on social media platforms for connectivity and other benefits. Unfortunately, fake news and misinformation regarding this virus is also available to people and causing some massive problems. So, fighting this infodemic has become a significant challenge. We present our solution for the "Constraint@AAAI2021 - COVID19 Fake News Detection in English" challenge in this work. After extensive experimentation with numerous architectures and techniques, we use eight different transformer-based pre-trained models with additional layers to construct a stacking ensemble classifier and fine-tuned them for our purpose. We achieved 0.979906542 accuracy, 0.979913119 precision, 0.979906542 recall, and 0.979907901 f1-score on the test dataset of the competition.
\end{abstract}
\section{Introduction}
The Coronavirus disease 2019 (COVID-19) is an infectious disease caused by SARS coronavirus 2. It has impacted almost every country and changed people worldwide's social, economic, and psychological states. Especially in the past few months, people have become more information-hungry on this topic. Hence, they are exposed to a significant amount of interaction to the information circled around coronavirus through various platforms. In parallel, the infodemic of false and misinformation regarding the virus has been rising. 

 As the idea of "stay-at-home" is proved to be the most effective precaution against the virus, people's most preferred solution for communication or entertainment has been various online and social media platforms. Hence, the number of people exposed to rumors or misinformations is more significant than ever before. In \cite{facebook_ad}, Facebook advertisements across 64 countries were examined, and it was found that 5\% of the advertisements contain possible errors or misinformations. Besides, many online news portals purposefully present misguiding news to increase their popularity. In recent times, a cluster of fake news about lock-downs, possible remedies, vaccinations have caused panic among people. People started to pile up stocks of sanitizers, masks, and the supply chain disrupted out of fear. Fighting against this ever-increasing amount of fake or misleading news has proven to be as crucial as finding remedies against the virus to overcome the pandemic.  

Fake news detection is a critical task as it requires the identification and detection of various news types like clickbait, propaganda, satire, misinformation, falsification, sloppy journalism, and many more. Traditionally, machine learning-based classifiers have been a go-to solution in this domain. In recent years, sequence models like RNN, LSTM, and CNN have shown good competency. However, the introduction of transformers has caused a considerable performance gain. In this work, we propose a solution to the "Constraint@AAAI2021 - COVID19 Fake News Detection in English" task on the provided dataset\cite{competition_dataset}. Our contributions to this task are the following:
\begin{itemize}
    \item We perform extensive experimentation like training classic machine learning models and traditional text classification models like Bidirectional LSTM, one dimensional CNN on the competition dataset, and incorporating publicly available external datasets for training our models.
    
    \item We fine-tune state-of-the-art transformer based pre-trained models as per the requirement of our task. We also experiment with additional R-CNN \cite{rcnn}, multichannel CNN  with attention \cite{amcnn}, multilayer perceptron (MLP) modules to increase the model's performance. Finally, we construct a stacking ensemble classifier of eight transformer models BERT\cite{bert}, GPT-2\cite{gpt-2}, XLNet\cite{xlnet}, RoBERTa\cite{roberta}, DistillRoBERTa, ALBERT\cite{albert}, BART\cite{bart}, and DeBERTa\cite{deberta}, each with additional MLP layers. 
    
    \item We also present our overall workflow, comparison, and performance analysis of our experiments on the validation set.
\end{itemize}

\section{Related Work}
The issue of fake news detection has been well studied in various fields. Existing machine learning-based methods like support vector machine (SVM)\cite{svm}, decision tree \cite{decision_tree} have worked as baselines for this task. In \cite {riedel_fake}, a simple classifier was used to leverage the Term Frequency(TF), Term Frequency and Inverse Document Frequency(TF-IDF), and cosine similarity between vectors as features to provide a baseline for fake news detection task on the Fake News Challenge (FNC-1) dataset\footnote{\url{http://www.fakenewschallenge.org/}}.

Deep learning linguistic models like Convolutional Neural Networks (CNN), Recurrent Neural Network (RNN), and Long Short-Term Memory (LSTM) can analyse variable length sequential data and discover hidden complex patterns in textual data. In \cite {bidirectional_example_1} , a fake news detection model based on Bidirectional LSTM\cite {bi-lstm} is presented. In \cite{semantic_matching}, Bidirectional LSTM along with GloVe\cite{glove} and ELMO\cite{elmo} embeddings were used to encode the claim and sentences.

Despite the efficacy of the above-stated language models, they still fell short of producing significant accuracy. In recent years, transformers and their various modifications have brought about considerable performance improvement in various natural language processing tasks. In \cite{exbake}, the contextual relationship between the headline and main text of news was analysed using BERT\cite{bert}. They showed that fine-tuning the pre-trained BERT for specific tasks like fake news detection can outperform existing sequence models. 

Research efforts focusing on COVID-19 fake news detection have also been made. In \cite{detecting_misleading_information_covid19}, a detection model based on the information provided by the World Health Organization, UNICEF, and the United Nations was proposed. Ten machine learning algorithms were used along with a voting ensemble classifier. In \cite{fakecovid}, a Twitter dataset containing 7623 tweets and corresponding labels was introduced. 

\section{Models}
We use an ensemble of multiple varieties of pre-trained transformers for the task at hand. Transformers are state-of-the-art tools for natural language processing. They have stacked blocks of identical encoders and decoders with self-attention. Following are the short descriptions of the varieties of transformers that are used: 
\begin{enumerate}
    \item \textbf{BERT\cite{bert}:} BERT(Bidirectional Encoder Representations from Transformers) is a bidirectional language transformer. It is trained based on masked language modeling and next sentence prediction on a sizeable unlabelled text corpus.
    
    \item \textbf{GPT-2\cite{gpt-2}:} It is a modified transformer with a larger context and vocabulary size. In contrast to the ordinary transformers, it has an additional normalization layer after the self-attention block. 
    
    \item \textbf{XLNet\cite{xlnet}:} It is a modified version of the transformer-XL. It is trained to learn the bidirectional context with an autoregressive method.
    
    \item \textbf{RoBERTa\cite{roberta}:} It is an optimized BERT. It is trained or more robust data and includes fine-tuning the original BERT without the next-sentence prediction objective.
    
    \item \textbf{DistilRoBERTa:} It is a distilled and faster version of the RoBERTa base version.
    
    \item \textbf{ALBERT\cite{albert}:} It splits the embedding matrix into two smaller metrics and uses two-parameter reduction techniques to increase training speed and decrease memory consumption. 
    
    \item \textbf{Bart\cite{bart}:} It is introduced by facebook as a sequence-to-sequence machine translation model. It is a fusion between BERT\cite{bert} and GPT\cite{gpt}. 
    
    \item \textbf{DeBERTa\cite{deberta}:} It is built on RoBERTa with some modifications like disentangled attention and enhanced mask decoder. 
\end{enumerate}

\textbf{Multilayer Perceptron (MLP):} Each of the previous models is followed by a multilayer perceptron module. It consists of a fully connected layer with 64 hidden units, followed by a normalization layer, followed by a linear layer with tanh activation, a dropout layer and a softmax layer for generating two class classification probabilities. All the weights are initialized by Xavier initialization. 

\begin{figure}
\begin{center}
\includegraphics[width=\textwidth]{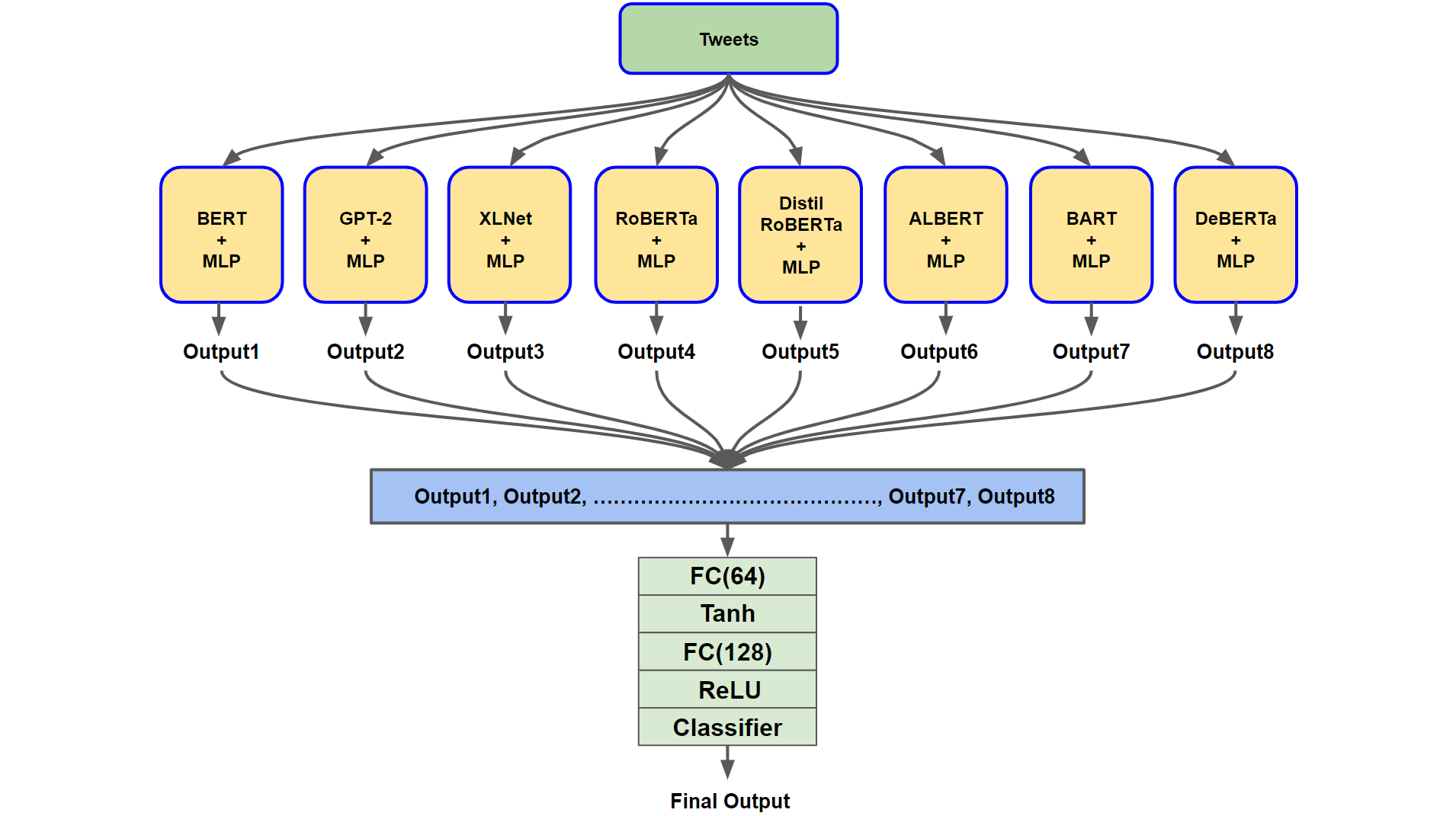}
\caption{Overview of our proposed method. tweets are given as inputs to individual models to produce outputs. then all the outputs are combined to build a 1x8 feature vector and are fed to the meta learner to generate the ensembled final output.} \label{fig1}
\end{center}
\end{figure}

\textbf{Ensemble Module:} For ensembling, we train a meta-learner consisting of the following units: a fully connected layer with 64 hidden units, a linear layer with tanh activation function, followed by a fully connected layer with 128 hidden units, a linear layer with ReLU activation function, and a softmax classifier. Figure \ref{fig1} contains an overview of our proposed architecture.

\section{Dataset}
The dataset\cite{competition_dataset} provided for the competition has social media posts related to COVID-19 and corresponding labels indicating if the social media posts are fake or real. The dataset is divided into three sets, which are train, validation, and test. The train set contains 6420 samples. Test and validation set each contains 2140 social media posts with their corresponding labels. The train set consists of 3360 real and 3060 fake social media posts, whereas the test and validation set consists of 1120 real and 1020 fake social media posts. On average, the train, validation, and test set contain 27.0, 26.79, and 27.46 words, respectively, in a social media post.
We also used additional data from \cite{fakecovid} for training our model as experimentation. It has 7623 labeled data. We used each data title as a replacement of 'tweet' in the competition data because both have similar lengths. We converted the multi-label classification data to binary as per the label type. As a result, 7501 were classified as fake, and 122 were classified as real.

\section{Experimentation and Result Analysis}
We perform experiments primarily on traditional language models such as Bidirectional LSTM(Bi-LSTM) \cite{bi-lstm} with attention, 1 dimensional CNN(1D-CNN), Hierarchical Attention Networks(HAN)\cite{han}, Recurrent Convolutional Neural Networks(RCNN)\cite{rcnn},  and Multichannel CNN with Attention(AMCNN)\cite{amcnn} on the competition dataset. We also experiment with transformer-based pre-trained models like BERT and RoBERTa. The result of these experiments is shown in the table \ref{tab:table1}.
\begin{table}[]
\caption{Comparison of the performance of transformer based models with traditional language models on the validation dataset}
\label{tab:table1}
\begin{center}
\begin{tabular}{|c|c|c|c|c|c|c|c|}
\hline
Model               & Accuracy & \multicolumn{2}{c|}{f1-score} & \multicolumn{2}{c|}{Precision} & \multicolumn{2}{c|}{Recall} \\ \hline
                    &          & Fake          & Real          & Fake           & Real          & Fake         & Real         \\ \hline
Bi-LSTM + attention   & 0.928      & 0.931     & 0.924       & 0.931      & 0.925     & 0.932   & 0.924   \\ \hline
1D-CNN         & 0.926      & 0.931     & 0.920       & 0.908      & 0.948     & 0.948   & 0.948   \\ \hline
HAN         & 0.930      & 0.933     & 0.928       & 0.943      & 0.918     & 0.923   & 0.938   \\ \hline
AMCNN         & 0.926      & 0.931     & 0.920       & 0.908      & 0.949     & 0.956   & 0.893   \\ \hline
RCNN         & 0.933      & 0.937     & 0.928       & 0.921      & 0.947     & 0.954   & 0.910   \\ \hline
BERT          & 0.971 & 0.969      & 0.972      & 0.977       & 0.965      & 0.961     & 0.980     \\ \hline
RoBERTa       & \textbf{0.979} & \textbf{0.977}      & \textbf{0.980}      & \textbf{0.981}       & \textbf{0.976}      & \textbf{0.974}     & \textbf{0.983}     \\ \hline
\end{tabular}
\end{center}
\end{table}

From table \ref{tab:table1}, it is clearly evident that transformer-based pre-trained models showed radical improvements on scores than the traditional models. But in these experiments, we use just a dense layer with a softmax classifier on top of the transformers for training and prediction. Still, there is a lot of scopes to improve it further. According to table \ref{tab:table1} it is also clear that, among  the traditional models scores for RCNN is better than the others. So, we choose RCNN and add it on top of BERT and RoBERTa and train these combined models. In these cases, we find that none of these additions improves the performance of the transformer-based models with a simple classification layer. The output of these experiments is shown in table \ref{tab:table2} with the corresponding model's name.
\begin{table}[]
\caption{Comparison of the performance among different  models combined with transformer based models on the validation dataset}
\label{tab:table2}
\begin{center}
\begin{tabular}{|c|c|c|c|c|c|c|c|}
\hline
Model               & Accuracy & \multicolumn{2}{c|}{f1-score} & \multicolumn{2}{c|}{Precision} & \multicolumn{2}{c|}{Recall} \\ \hline
                    &          & Fake          & Real          & Fake           & Real          & Fake         & Real         \\ \hline
BERT + RCNN         & 0.967 & 0.965      & 0.969      & 0.980       & 0.956      & 0.950    & 0.982     \\ \hline
RoBERTa + RCNN  & 0.968 & 0.966  & 0.970  & \textbf{0.988}   & 0.956    & 0.950   & \textbf{0.989}    \\ \hline
RoBERTa + SVM       & 0.978 & 0.977      & 0.979      & 0.987       & 0.970      & 0.967     & 0.988     \\ \hline
RoBERTa + MLP       & \textbf{0.979} & \textbf{0.977}      & \textbf{0.980}     & 0.981          & \textbf{0.976}      & \textbf{0.974}    & 0.983    \\ \hline
\end{tabular}
\end{center}
\end{table}

We experiment with a few classic machine learning classifiers such as Decision Tree(DT)\cite{decision_tree}, support vector machine (SVM)\cite{svm} with different kernels on top of transformer-based models. Among these, SVM with Radial Basis Function(RBF) kernel achieves highest score which is almost similar to the simple classifier with a slight improvement. We also experiment with different combinations of MLP on top of RoBERTa and find that the best performing MLP performed better than SVM with RBF kernel. The results are shown in table \ref{tab:table2}.


Since none of those models except MLP on top of RoBERTa  perform better than the transformer-based models with a simple classifier, we add a Multi-Layer Perceptron (MLP) on top of transformer-based models. We test different combinations of MLPs and choose the best performing combination among these. 

Initially, we experiment with only BERT and RoBERTa among the transformer-based pre-trained models. To add diversity and capture different hidden information of the data, we select the most suitable eight models among different transformer-based models based on their performance and low resource necessity. These are \textbf{BERT}, \textbf{RoBERTa}, \textbf{XLNet}, \textbf{GPT-2}, \textbf{ALBERT}, \textbf{DistilRoBERTa}, \textbf{BART} and \textbf{DeBERTa}. We experiment on each of these models with base architecture since we do not have enough resources. We add the best performing MLP on top of each. These models are tuned individually on the train and validation dataset. The results of the individual performance are shown in table \ref{tab:table4}. 

\begin{table}[]
\caption{Comparison of the performance of various modified transformer models on the validation dataset}
\label{tab:table4}
\begin{center}
\begin{tabular}{|c|c|c|c|c|c|c|c|}
\hline
Model               & Accuracy & \multicolumn{2}{c|}{f1-score} & \multicolumn{2}{c|}{Precision} & \multicolumn{2}{c|}{Recall} \\ \hline
                    &          & Fake          & Real          & Fake           & Real          & Fake         & Real         \\ \hline
RoBERTa + MLP   & \textbf{0.979} & \textbf{0.977}      & \textbf{0.980}      & 0.981       & \textbf{0.976}      & \textbf{0.974}     & 0.983     \\ \hline
BERT + MLP          & 0.971 & 0.969      & 0.972      & 0.977       & 0.965      & 0.961     & 0.980     \\ \hline
XLNet + MLP   & 0.977 & 0.975     & 0.978      & 0.982       & 0.972      & 0.969     & 0.984     \\ \hline
GPT-2 + MLP   & 0.974 & 0.973      & 0.976      & 0.974       & 0.974      & 0.972     & 0.977     \\ \hline
DistilRoBERTa + MLP & 0.978 & 0.977      & 0.979      & 0.986       & 0.971      & 0.968     & 0.988     \\ \hline
ALBERT + MLP        & 0.962 & 0.960      & 0.964      & 0.976       & 0.951      & 0.944     & 0.979     \\ \hline
BART + MLP  & 0.978  & 0.976      & 0.979      & \textbf{0.988}       & 0.969      & 0.965     & \textbf{0.989}     \\ \hline
DeBERTa + MLP  & 0.973 & 0.971      & 0.975      & 0.988       & 0.960      & 0.955     & \textbf{0.989}     \\ \hline
\end{tabular}
\end{center}
\end{table}

From table \ref{tab:table4}, it can be carefully observed that different models among the transformer-based models with MLP can capture features differently. As a result, their combination in prediction can help to improve the overall performance. So, we finally experiment with different ensemble techniques.

For ensembling, we build a dataset based on the predictions from the individual models of table \ref{tab:table4}. We form a feature vector where the features are predictions from each previous model for a particular training sample. We use it to train a meta-learner to model the individual predictions into a more generalized final output. We experiment with random forest and SVM classifiers and achieve accuracy of .9785 and .9789, respectively on the validation set. Since these methods do not yield significant improvement, we introduce a more robust and complex meta-learner consisting of fully connected layers and achieve considerable performance gain. We try different combinations of ensemble of transformer based models also. The results of best 3 ensembled models are summarised in table \ref{tab:ensemble_table}.

\begin{table}[]
\caption{Quantitative comparison of ensemble models on the validation dataset. }
\label{tab:ensemble_table}
\begin{center}
\begin{tabular}{|c|c|c|c|c|c|c|c|}
\hline
Model                & Accuracy        & \multicolumn{2}{c|}{f1-score}     & \multicolumn{2}{c|}{Precision}    & \multicolumn{2}{c|}{Recall}       \\ \hline
                     &                 & Fake            & Real            & Fake            & Real            & Fake            & Real            \\ \hline
Ensemble-v1 & 0.981          & 0.980          & 0.982          & 0.980          & 0.983          & 0.981          & 0.981          \\ \hline
Ensemble-v2 & 0.983          & 0.982          & 0.984          & \textbf{0.986} & 0.980          & 0.978          & \textbf{0.988} \\ \hline
Ensemble-v3 & \textbf{0.984} & \textbf{0.983} & \textbf{0.984} & 0.983          & \textbf{0.984} & \textbf{0.982} & 0.985          \\ \hline
\end{tabular}
\end{center}
\end{table}

Here, Ensemble-v1 consists of RoBERTa, BERT, XLNet, GPT-2. Ensemble-v2 has ALBERT, BART, DeBERTa, and the previous models from Ensemble-v1. DistilRoBERTa is added to the previous seven models in the Ensemble-v3. From table \ref{tab:ensemble_table}, it can be seen that having more individual models in the final ensemble classifier usually yields better generalisation and achieves higher accuracy. 

One crucial factor is that despite adding additional data for training, all of our models' performance in the validation set decreases. One probable reason for this decline might be the sizeable imbalance in the fake and real news counts. So, we choose to discard additional data for our final model training.

\subsection{Hyper-parameters}
We test different hyper-parameters like the number of layers, number of units in a layer, learning rate, weight decay, dropouts, normalization, etc. within a feasible range. In all the models we use different learning rates between \textbf{1e-3}  and \textbf{2e-6}. Learning rate \textbf{2e-6} is used in most cases since the dataset is not so large. Using this learning rate helps to converge to the global minimum easily. For traditional models, we test a different combination of several layers with many units. We find Bidirectional LSTM models with 256 layers with 128 hidden units performed best. For 1D-CNN models, we use a layer with 256 filters and filter sizes 1-6. For all traditional models, we use a similar structure with a little variation. In case of the transformer-based pre-trained models, we use the base architectures. It is because using large models on small datasets can cause the models to over-fit. Also, we face a resource limitation for experimenting with larger models.

\section{Conclusion}
In this work, we have presented our overall workflow for the fake news detection task. We have conducted a number of experiments and provided a comprehensive solution based on modified transformers with additional layers and an ensemble classifier. Our method achieves comparative accuracy on the test dataset and therefore helps us to place 20 in the leaderboard of "CONSTRAINT 2021 Shared Tasks: Detecting English COVID-19 Fake News and Hindi Hostile Posts"\cite{taskoverview} competition. Based on our experience from this task, we feel that a more balanced additional training data may help our model perform better. Besides, extracting more information like parts of speech, named entities, number of words, punctuation, hashtags, website links etc. in social media posts and using these as meta-data during training can increase the methodology's performance. 

\bibliographystyle{splncs04}
\bibliography{main}

\begin{thebibliography}{10}
\providecommand{\url}[1]{\texttt{#1}}
\providecommand{\urlprefix}{URL }
\providecommand{\doi}[1]{https://doi.org/#1}

\bibitem{bidirectional_example_1}
Bahad, P., Saxena, P., Kamal, R.: Fake news detection using bi-directional
  lstm-recurrent neural network. Procedia Computer Science  \textbf{165},
  74--82 (2019)

\bibitem{decision_tree}
Breiman, L., Friedman, J., Olshen, R., Stone, C.: Classification and regression
  trees (wadsworth, belmont, ca). ISBN-13 pp. 978--0412048418 (1984)

\bibitem{bert}
Devlin, J., Chang, M.W., Lee, K., Toutanova, K.: Bert: Pre-training of deep
  bidirectional transformers for language understanding. arXiv preprint
  arXiv:1810.04805  (2018)

\bibitem{detecting_misleading_information_covid19}
Elhadad, M.K., Li, K.F., Gebali, F.: Detecting misleading information on
  covid-19. IEEE Access  \textbf{8},  165201--165215 (2020)

\bibitem{svm}
Girosi, F., Niyogi, P., Poggio, T., Vapnik, V.: Comparing support vector
  machines with gaussian kernels to radial basis function classifiers. Tech.
  rep., Technical Report 1599, Massachusetts Institute of Techology, MA, USA
  (1996)

\bibitem{deberta}
He, P., Liu, X., Gao, J., Chen, W.: Deberta: Decoding-enhanced bert with
  disentangled attention. arXiv preprint arXiv:2006.03654  (2020)

\bibitem{exbake}
Jwa, H., Oh, D., Park, K., Kang, J.M., Lim, H.: exbake: Automatic fake news
  detection model based on bidirectional encoder representations from
  transformers (bert). Applied Sciences  \textbf{9}(19), ~4062 (2019)

\bibitem{rcnn}
Lai, S., Xu, L., Liu, K., Zhao, J.: Recurrent convolutional neural networks for
  text classification. In: Twenty-ninth AAAI conference on artificial
  intelligence (2015)

\bibitem{albert}
Lan, Z., Chen, M., Goodman, S., Gimpel, K., Sharma, P., Soricut, R.: Albert: A
  lite bert for self-supervised learning of language representations. arXiv
  preprint arXiv:1909.11942  (2019)

\bibitem{bart}
Lewis, M., Liu, Y., Goyal, N., Ghazvininejad, M., Mohamed, A., Levy, O.,
  Stoyanov, V., Zettlemoyer, L.: Bart: Denoising sequence-to-sequence
  pre-training for natural language generation, translation, and comprehension.
  arXiv preprint arXiv:1910.13461  (2019)

\bibitem{roberta}
Liu, Y., Ott, M., Goyal, N., Du, J., Joshi, M., Chen, D., Levy, O., Lewis, M.,
  Zettlemoyer, L., Stoyanov, V.: Roberta: A robustly optimized bert pretraining
  approach. arXiv preprint arXiv:1907.11692  (2019)

\bibitem{amcnn}
Liu, Z., Huang, H., Lu, C., Lyu, S.: Multichannel cnn with attention for text
  classification. arXiv preprint arXiv:2006.16174  (2020)

\bibitem{facebook_ad}
Mejova, Y., Weber, I., Fernandez-Luque, L.: Online health monitoring using
  facebook advertisement audience estimates in the united states: evaluation
  study. JMIR public health and surveillance  \textbf{4}(1), ~e30 (2018)

\bibitem{semantic_matching}
Nie, Y., Chen, H., Bansal, M.: Combining fact extraction and verification with
  neural semantic matching networks. In: Proceedings of the AAAI Conference on
  Artificial Intelligence. vol.~33, pp. 6859--6866 (2019)

\bibitem{taskoverview}
Patwa, P., Bhardwaj, M., Guptha, V., Kumari, G., Sharma, S., PYKL, S., Das, A.,
  Ekbal, A., Akhtar, S., Chakraborty, T.: Overview of constraint 2021 shared
  tasks: Detecting english covid-19 fake news and hindi hostile posts. In:
  Proceedings of the First Workshop on Combating Online Hostile Posts in
  Regional Languages during Emergency Situation ({CONSTRAINT}). Springer (2021)

\bibitem{competition_dataset}
Patwa, P., Sharma, S., PYKL, S., Guptha, V., Kumari, G., Akhtar, M.S., Ekbal,
  A., Das, A., Chakraborty, T.: Fighting an infodemic: Covid-19 fake news
  dataset. arXiv preprint arXiv:2011.03327  (2020)

\bibitem{glove}
Pennington, J., Socher, R., Manning, C.D.: Glove: Global vectors for word
  representation. In: Empirical Methods in Natural Language Processing (EMNLP).
  pp. 1532--1543 (2014), \url{http://www.aclweb.org/anthology/D14-1162}

\bibitem{elmo}
Peters, M.E., Neumann, M., Iyyer, M., Gardner, M., Clark, C., Lee, K.,
  Zettlemoyer, L.: Deep contextualized word representations. arXiv preprint
  arXiv:1802.05365  (2018)

\bibitem{gpt}
Radford, A., Narasimhan, K., Salimans, T., Sutskever, I.: Improving language
  understanding by generative pre-training (2018)

\bibitem{gpt-2}
Radford, A., Wu, J., Child, R., Luan, D., Amodei, D., Sutskever, I.: Language
  models are unsupervised multitask learners. OpenAI blog  \textbf{1}(8), ~9
  (2019)

\bibitem{riedel_fake}
Riedel, B., Augenstein, I., Spithourakis, G.P., Riedel, S.: A simple but
  tough-to-beat baseline for the fake news challenge stance detection task.
  arXiv preprint arXiv:1707.03264  (2017)

\bibitem{bi-lstm}
Schuster, M., Paliwal, K.K.: Bidirectional recurrent neural networks. IEEE
  transactions on Signal Processing  \textbf{45}(11),  2673--2681 (1997)

\bibitem{fakecovid}
Shahi, G.K., Nandini, D.: Fake{C}ovid -- a multilingual cross-domain fact check
  news dataset for covid-19. In: Workshop Proceedings of the 14th International
  {AAAI} {C}onference on {W}eb and {S}ocial {M}edia (2020),
  \url{http://workshop-proceedings.icwsm.org/pdf/2020_14.pdf}

\bibitem{xlnet}
Yang, Z., Dai, Z., Yang, Y., Carbonell, J., Salakhutdinov, R.R., Le, Q.V.:
  Xlnet: Generalized autoregressive pretraining for language understanding. In:
  Advances in neural information processing systems. pp. 5753--5763 (2019)

\bibitem{han}
Yang, Z., Yang, D., Dyer, C., He, X., Smola, A., Hovy, E.: Hierarchical
  attention networks for document classification. In: Proceedings of the 2016
  conference of the North American chapter of the association for computational
  linguistics: human language technologies. pp. 1480--1489 (2016)

\end{thebibliography}
\end{document}